# AGISim, An Open Source Airborne Gimbal Mounted IMU Signal Simulator Considering Flight Dynamics Model

Alireza Kazemi [1,3*], Reza Rohani Sarvestani [2,3]
1-Assitant Prof., Mathematics Dept., Faculty of Science, Shiraz University, Fars, Iran, alireza.kazemi@shirazu.ac.ir
2-Assistant Prof., Computer Eng. Dept., Engineering School, Shahrekord University, Iran, rrohani@sku.ac.ir
3- Research Dept., System Intelligizers Co. (SICO), Shiraz, Fars, Iran
*Corresponding Author

**Abstract**
In this work we present more comprehensive evaluations on our airborne Gimbal mounted inertial measurement unit (IMU) signal simulator which also considers flight dynamic model (FDM). A flexible IMU signal simulator is an enabling tool in design, development, improvement, test and verification of aided inertial navigation systems (INS). Efforts by other researchers had been concentrated on simulation of the strapdown INS (SINS) with the IMU rigidly attached to the moving body frame. However custom airborne surveying/mapping applications that need pointing and stabilizing camera or any other surveying sensor, require mounting the IMU beside the sensor on a Gimbal onboard the airframe. Hence the proposed Gimbal mounted IMU signal simulator is of interest whilst itself requires further analysis and verifications. Extended evaluation results in terms of both unit tests and functional/integration tests (using aided inertial navigation algorithms with variable/dynamic lever arms), verifies the simulator and its applicability for the mentioned tasks. We have further packaged and published our MATLAB code for the proposed simulator as an open source GitHub repository[1].

## 1. Introduction

Aided inertial navigation systems have essential role in navigation, surveillance, control and localization of moving bodies specifically for airborne platforms. It is also a major building block in motion compensation, mosaicking, merging and geo-spatial tagging of the observations captured by the surveying sensors being onboard airborne platforms. Aided INS provides temporal estimates of position, velocity and attitude in 3D space and is composed of three major devices [1] [2] [3]:

1. **IMU**: inertial measurement unit sensor provides the observed average acceleration ($\Delta v$) and rotation rate ($\Delta \theta$) in 3D Cartesian frame around its body.
2. **Aiding sensor**: a sensor such as satellite navigation (GNSS), vehicle odometers [4], tachometers [5], LiDAR [6], barometer and optical flow sensors in UAVs [7] or any sensor type providing periodic velocity and or position measures to fix the INS estimated navigation solution which will otherwise suffer from ever increasing velocity/position drift. The non-linear position drift is due to mixing and twice integration of the noisy IMU ($\Delta v, \Delta \theta$) measures affected by IMU biases, scale factors, random walk, thermal and quantization noises.
3. **Navigation processor**: It applies INS mechanization equations to the IMU data which adjusts and then integrates acceleration and rotation rates to estimate velocity, position and attitude and further refine the estimation by fusing/fixing it with the data coming from aiding sensor. This is usually done using kind of a recursive estimator such as Kalman filter (KF).

Historically the inertial navigation systems were initially developed based on gimbaled INS rather than strapdown INS [8] in 60s decade because of the lack of gyro sensor technology and digital computers. Such a system needed an accurate and expensive mechanical gimbal to suspend and stabilize the accelerometer sensors apparatus. After development of gyro sensor and digital computers, the strapdown INS method was preferred to save weight, volume, complexity and price and yet provide good accuracies [9].

It is important to mention that the term *gimbal mounted*, in this paper does not necessarily refer to that historical gimbaled INS. Since our method can simulate IMU accelerometer and gyro signals of a strapdown INS system mounted on a gimbal, or it may be adapted to just simulate the accelerometer data of a historical gimbaled INS systems. This also benefits in conducing comparative studies. The current role of gimbal if it is present is to point its surveying sensor(s) in a stabilized direction or fix at a desired ground point and it is not to suspend IMU unit as was in the gimbaled INS method. The reason for mounting IMU on the gimbal in strapdown systems is two ply:

1. To provide the IMU's gyro raw angular rates to the stabilization/pointing control loop of the gimbal.
2. To be used to estimate accurate position, velocity and attitude of the surveying sensor which in some cases is required for motion compensation [10] and or accurate geo/spatial tagging of the observations captured from the surveying sensors.

---
[1] https://github.com/sico-res/ag-imu-sim

This custom mounting of IMU also requires custom aided INS integration algorithms in which the variable/dynamic lever arm caused by relative (non-rigid) motion between observation center of IMU and aiding sensor is accounted for [11] [12].

Design, development, improvement, and verification of (custom) integrated inertial navigation systems requires testing the performance of algorithms in many diverse scenarios in which sensor measures or integration algorithm may present different error kinds or levels due to specific platform situations [5]. Trying to tackle such an exhaustive test scenario using real sensors and real moving platforms specifically for airborne platform (that needs flight), imposes high costs and complex challenges like acquiring aircraft, flight crew, flight permissions and suitable whether [3]. A well designed IMU signal simulator together with proper aiding sensors simulations helps much to overcome such challenges, alleviate costs and make feasible the thorough test and evaluation of custom integrated navigation systems.

### 1.1. Related Work

Although IMU signal simulation research has been subject of attention in the navigation community especially since 2007 [13], but a gimbal mounted strapdown IMU simulator, onboard an airborne platform considering both flight dynamics model (FDM) and gimbal dynamics model was not available. In [14] and [16] IMU signal simulation by analyzing software architectural and error modeling is presented. A stepwise description of IMU signal simulation using MATLAB code is provided in [15] in which input platform motion is considered to be simple linear or circular paths. Groves has provided a good open source IMU simulator as companion materials of his comprehensive book on the principles of INS [1], but it also lacks realistic platform FDM. In [3] an open source IMU signal simulator based on Grove's MATLAB codes is presented that also benefits from realistic FDM of the open source C++ JSBSim [17] flight simulator as the input motion stimuli for IMU. Some practical considerations on IMU data simulation is provided in [18]. In [5] a custom realistic 9DoF IMU and tachometer signal simulation for evaluation and test of train safety navigation systems is presented that uses railway digital maps as input motion stimuli. A simulation platform for an integrated navigation algorithm for hypersonic vehicles based on flight mechanics is proposed in [19] where the generation method of inertial measurement unit data and satellite receiver data is introduced. A simulator for onboard sensors of a quadrotor platform including IMU sensor is also developed in [7].

None of the above simulators considers mounting the IMU on a gimbal. This is very important in accurate measure of the motion (and potentially higher frequency vibrations) of the surveying sensors mounted on a 3DoF gimbal [10]. Such a gimbal is considered for fixed pointing or stabilizing the surveying sensor(s). However the pointing/stabilizing control loop cannot completely tackle the stabilization and hence accurate motion measurement is required to estimate and compensate for the stabilization/pointing residuals/errors.

In [20] we have proposed an airborne gimbal mounted IMU signal simulator that jointly adopts flight dynamics models and considers gimbal motion/rotation effects in the generated IMU signals. The previous brief report of our flexible simulator is now extended with more comprehensive evaluation results and is presented here. We have further packaged and published the MATLAB code for the proposed simulator as an open source GitHub repository [21].

### 1.2. Notation

Navigation analysis and modeling, generally involves equations with elaborate notation including multiple reference frames (*i.e.* IMU, gimbal, body, nav, inertial, ecef), dimensions (*i.e.* 3D-spatial, 1D-temporal), quantities (*i.e.* position, velocity, acceleration, attitude, angular velocity) and representation types (*i.e.* scalar, vector, matrix, quaternion). Furthermore it is very common to have symbols with three super- and sub-scripts (*e.g.* $\boldsymbol{f}_{\mathrm{ib_I}}^{\mathrm{e}}$) for quantities (*i.e.* $\boldsymbol{f}$) measured between two coordinate frames (*i.e.* i, $\mathrm{b_I}$) which must be resolved in a third frame (*i.e.* e). Hence we specify clear notation rules to increase readability.

As the general notation, we have used non-italic letters (*e.g.* x, X) for scalar constants and object labels (*i.e.* frames, axes and points names), italic-lower-case letters (*e.g.* $x$) for scalar variables, italic-bold-lower-case letters (*e.g.* $\boldsymbol{x}$) for vectors and italic upper-case letters (*e.g.* $X$) for matrices. Some more details on notation is presented in Table 1.

Table 1. Notation and symbols

| Type | Notation | Sym Group | Symbols |
|---|---|---|---|
| Constants | non-italic (*e.g.* x, X) | frame labels | $\mathrm{b_I}$(IMU)  $\mathrm{b_c}$(body)  $\mathrm{g_P}$(pan)  $\mathrm{g_T}$(tilt)  $\mathrm{g_R}$(roll)  n(nav)  i(inertial)  e(ecef) |
| | | axes labels | P, T, R (pan-tilt-roll)  X, Y, Z (cartesian XYZ)  N, E, D (cartesian NED) |
| Scalar variable | italic-lower-case (*e.g.* $x$) | time | $t$ |
| | | lever arms | $l_{\mathrm{PT}}$ (pan-tilt arm length)  $l_{\mathrm{TR}}$ (tilt-roll arm length) |
| Vector variable | italic-bold-lower-case (*e.g.* $\boldsymbol{x}$) | pose | $\boldsymbol{p}$(pos)  $\boldsymbol{v}$(vel)  $\boldsymbol{\Theta}$(att) |
| | | true inertial signals | $\boldsymbol{f}$(sp-force)  $\boldsymbol{\omega}$(rot-rate) |
| | | IMU signals | $\tilde{\boldsymbol{f}}$(sp-force)  $\tilde{\boldsymbol{\omega}}$(rot-rate) |
| | | IMU biases | $\boldsymbol{b}_\mathrm{a}$(ac-bias)  $\boldsymbol{b}_\mathrm{g}$(gy-bias) |
| | | IMU noise | $\boldsymbol{w}_\mathrm{a}$(ac-noise)  $\boldsymbol{w}_\mathrm{g}$(gy-noise) |
| Matrix variable | non-italic-upper-case (*e.g.* $X$) | DCMs | $C$ (direction cosine mat) |
| | | IMU scale factors | $M_\mathrm{a}$(accel-sf)  $M_\mathrm{g}$(gyro-sf) |

The rest of this paper is organized as follows. In section 2 the necessary background is reviewed and some aspects if the simulation problem is specified. In section 3 the generative equations for the proposed airborne and gimbal mounted IMU simulator and its components is discussed. In section 4, implementation of the real time simulation, numerical results and validation procedure is presented. Finally the work is concluded in section 5 where the future work that it make viable is also mentioned.

## 2. Problem Specification & Background

We aim to simulate velocity and angular increment ($\Delta v, \Delta \Theta$) signals of an IMU which is mounted on a 3 axis (3DoF) gimbal and the gimbal itself is mounted on an airborne platform. The flight dynamics and gimbal rotation dynamics must be incorporated in IMU signal simulation. Beside the simulated signals, the simulator also provides ground truth position, velocity and attitude data which makes it very helpful in development and validation of variable lever arm, airborne aided SINS processing algorithms with gimbal mounted IMU. The block diagram of a typical aided SINS system without considering the problem of variable/dynamic lever arms is shown in Fig. 1. This system is not applicable when the IMU or the aiding sensor of the system is mounted on a moving sub-platform such as gimbal in the case of the proposed IMU simulator. In that case SINS algorithms that incorporate variable or dynamic lever arm compensation e.g. [11] must be adopted.

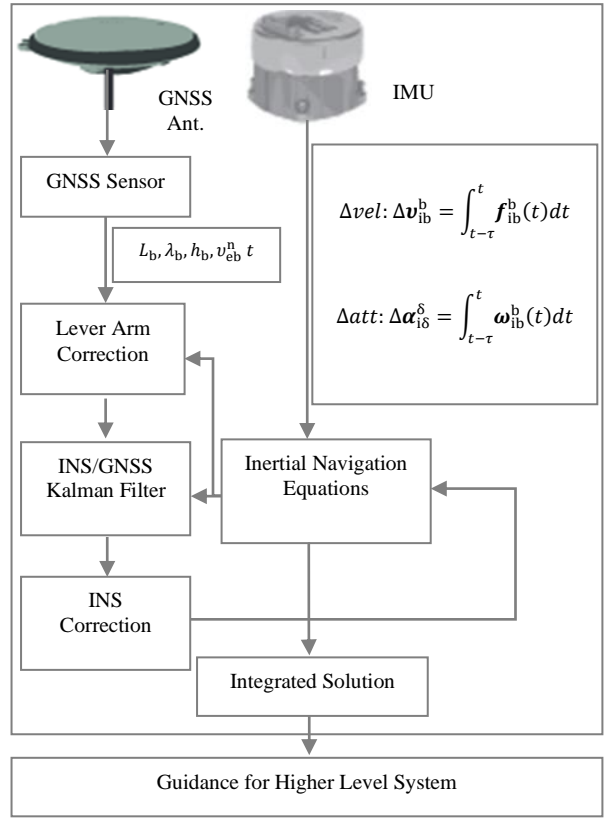

**Fig. 1.** Block diagram of a typical aided SINS system [3]

Since the focus of current work is on the design and validation of the gimbal mounted IMU simulator, we intentionally avoid mixing this work with the concerns of development or supplying of a variable lever arm SINS processing algorithm which itself requires a separate research task. Instead, we have designed a simplified yet adequate SINS processing system depicted in Fig. 2, which allows end to end verification of our gimbal mounted IMU simulator. Beside that this is not the only method for review and validation of the simulated data, since we have also unit test the simulator using intermediate data visualization and verification steps.

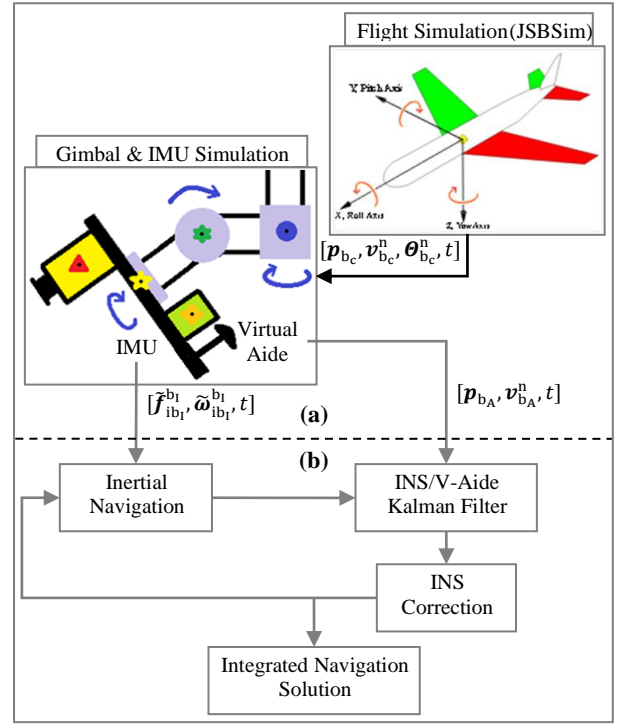

**Fig. 2.** Block diagram of the proposed system.
**(a)** the proposed simulator **(b)** the test suit for its functional/integration test

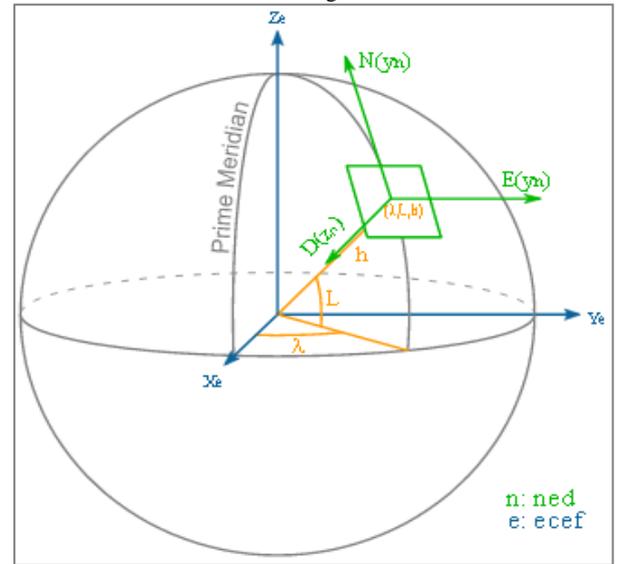

(a)

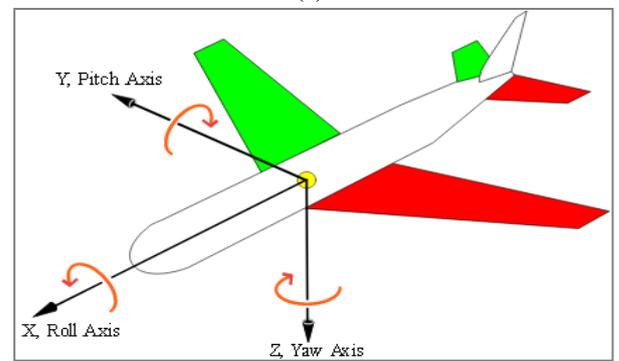

(b)

**Fig. 3.** Overview of frames and notations used in navigation
**(a)** Earth Center Earth Fixed (e), Local NED (n) and the background Inertial (i) frames **(b)** Platform body frame ($b_c$)
* Base SVG figures from Wikipedia, edited and re-designed by authors to present detailed concepts

In Fig. 2, we have devised a virtual/simulated aiding sensor beside the IMU, where measuring center of both is rigidly posed on the inner-plate frame of the gimbal. This configuration is used to simplify the evaluation procedure of the proposed IMU simulator by running a readily available, loosely coupled aided strapdown INS algorithm [1] [3], on the simulated signals without the need for dynamic lever arm compensation. The virtual aiding sensor is considered to be the same as a simulated GPS/GNSS receiver in terms of data type and accuracies except for the fact that the GNSS antenna (as its measuring center) cannot practically receive positioning signals if it is mounted on the gimbal which commonly resides in door or in a pod under the body of the airborne platform. In real world the GNSS receiver antenna is mounted on ceiling of aircraft.

The definition and notation of geo and platform reference frames presented in Fig. 3 are frequently referred in the navigation sensors and systems contexts. Hence we have illustrated it as the background and notation definition for the remaining content of the paper.

## 3. Airborne and Gimbal Mounted IMU Simulator

The proposed IMU signal simulator, requires specification of the following major components:

1. *Platform flight dynamics model (FDM)*: to generate ground truth position, velocity and attitude ($p_{b_c}^n$, $v_{b_c}^n$, $\Theta_{b_c}^n$) of the airborne platform body frame ($b_c$) relative to the local NED frame (n).
2. *Gimbal model*: to convert the ground truth platform body pose ($p_{b_c}$, $v_{b_c}^n$, $\Theta_{b_c}^n$) stage by stage into the IMU body frame ($p_{b_I}$, $v_{b_I}^n$, $\Theta_{b_I}^n$).
3. *Kinematic equations*: to generate the ground truth average specific force ($f_{ib_I}^{b_I}$) and rotation rates ($\omega_{ib_I}^{b_I}$) of IMU body frame ($b_I$) with respect to the inertial frame (i) measured in the IMU body frame.
4. *IMU error model*: to impose deterministic and stochastic disturbing factors on the ground truth specific force and rotation rates ($f_{ib_I}^{b_I}$, $\omega_{ib_I}^{b_I}$) to model the final IMU signals ($\tilde{f}_{ib_I}^{b_I}$, $\tilde{\omega}_{ib_I}^{b_I}$).

Each of the above components is detailed in the following.

### 3.1. Platform Flight Dynamics Model:

The base platform motion simulation is the initial stimuli to the IMU and generally any other motion measuring sensor. Both simplistic (e.g linear, piecewise linear, circular motion models [15]) or realistic (e.g. using flight dynamics [3] and train dynamics [5]) approaches are used in the literature to simulate platform motions. Since we have considered airborne platform, (as in [3]), to incorporate effects of realistic FDMs in the simulated IMU signals, we have adopted the open source C++ JSBSim flight dynamics simulator [17] to generate sequences of position, velocity and attitude for the aircraft platform in real time. In JSBSim, the aircraft parameters and flight path waypoints can be flexibly specified using XML configuration files and more importantly it provides an autopilot feature that can fly the aircraft on the specified scenario. Additionally the data logging format, sampling frequency and format of logging (offline in file vs. online on the network using different protocols and packet formats) may be specified via another configuration file.

We have used the same configuration as in open source project provided by [3] in which the time stamp, geodetic position (LLA), velocity and Euler angles (resolved in local NED frame) is logged at 50Hz on a user datagram protocol (UDP) port. These data are read from a UDP socket in MATLAB and used as simulated airborne platform motion profiles. It is worth to mention that JSBSim (like XPlane) is one of the most realistic real time FDM simulators available in the world.

### 3.2. Gimbal Modeling

We have modeled the 3DoF gimbal (see Fig. 4) by three perpendicular rotating junctions pan, tilt and roll, around the origins $o_{g_P}, o_{g_T}, o_{g_R}$ of the corresponding defined gimbal frames $g_P, g_T, g_R$ with possibly non-zero length $l_{PT}$ and $l_{TR}$ lever arms between junctions. Each of the gimbal frames $g_P, g_T, g_R$ rotates with a relative pan, tilt or roll (PTR) angle with respect to its previous base frame. Hence PTR angles must be applied different from Euler angles roll, pitch, yaw (RPY) of *e.g.* platform body frame in which RPY angles show rotations from the same base frame that is commonly either n (ned) or e (ecef).

Contrary to [11], the non-zero length lever arms $l_{PT}$ and $l_{TR}$ are considered between pan-tilt and tilt-roll junctions to generalize the model. This may be decided to be an over-design since gimbals are commonly designed to approximately be nested co-centric rings or hemispheres and thus their pan-tilt and tilt-roll arms are approximately zero. However for the gimbals which are designed to stabilize or point mounted sensors, due to volume considerations it is common that those lever arms have small non-zero lengths.

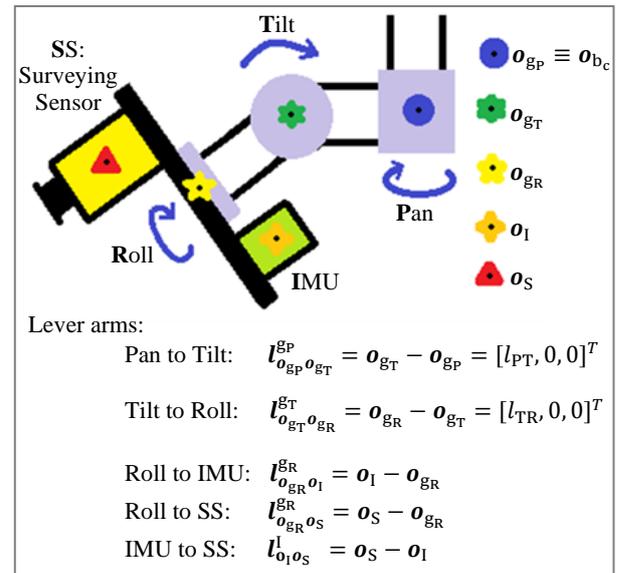

Fig. 4. Gimbal geometric model, showing origin of frames for three gimbal junctions, IMU and surveying sensor and the lever arm between these frames

In order to simulate $\Delta v$, $\Delta \Theta$ signals of the gimbal mounted IMU, first the ground-truth position, velocity and attitude (PVA) ($p_{b_I}$, $v_{b_I}^n$, $\Theta_{b_I}^n$) of the IMU body frame must be simulated. For that the PVA $p_{b_c}$, $v_{b_c}^n$, $\Theta_{b_c}^n$ of platform body frame, simulated by the JSBSim, must be transferred into the IMU body frame. This requires applying multiple frame to frame transposition steps.

Transposition of attitude just requires the static rotation of the source frames $b_1$ with respect to the reference frame (e.g. e or n) specified in the form of cosine transform matrix (CTM) $C_{b_1}^e$ and the relative rotation between the source and destination frames $C_{b_2}^{b_1}$.

$$C_{b_2}^n = C_{b_1}^n C_{b_2}^{b_1} \quad (1)$$

For transposition of the position, specified in geo coordinates latitude, longitude and altitude (LLA) as $p_{b_1} = [L, \lambda, h]$, the lever arm vector is also necessary and for transposition of velocity, specified in local ned frame n as $v_{b_1}^n$, additionally the rotation rate of the source frame $b_1$ with respect to the ecef frame e is needed. Following equations describe transposition of position and velocity between the two generic frames $b_1$ and $b_2$:

$$p_{b_2} = p_{b_1} + M_p C_{b_1}^n l_{b_1 b_2}^{b_1} \quad (2)$$

$$v_{b_2}^n = v_{b_1}^n + C_{b_1}^n \left( \omega_{eb_1}^{b_1} \times l_{b_1 b_2}^{b_1} \right) \quad (3)$$

$$M_p = \begin{bmatrix} 1/(R_N + h) & 0 & 0 \\ 0 & 1/((R_E + h) * \cos L) & 0 \\ 0 & 0 & -1 \end{bmatrix} \quad (4)$$

In this equations, $M_p$ is the Jacobean matrix for local approximation of coordinate transformation from the local Cartesian ned frame n into the LLA coordinates in which $R_N$ and $R_E$ are local meridian and transverse radius and $L$, $h$ are latitude and altitude of the origin of the source frame $b_1$, $l_{b_1 b_2}^{b_1}$ is the lever arm vector pointing from $b_1$ to $b_2$ resolved in the $b_1$ frame, $\times$ is the cross product operator and $\omega_{eb_1}^{b_1}$ is the rotation rate vector of frame $b_1$ with respect to the ecef fame e resolved in $b_1$. This rotation rate is computed from memorized CTM of the previous step $C_{b_1}^e(t - \Delta t)$ and that of the current step $C_{b_1}^e(t)$ as follows:

$$\Delta C_{b_1}^e = C_{b_1}^{e\ T}(t)\ C_{b_1}^e(t - \Delta t) \quad (5)$$

$$\omega_{eb_1}^{b_1} = \frac{1}{2\Delta t} \begin{bmatrix} \Delta C_{b_1}^e(2,3) - \Delta C_{b_1}^e(3,2) \\ \Delta C_{b_1}^e(3,1) - \Delta C_{b_1}^e(1,3) \\ \Delta C_{b_1}^e(1,2) - \Delta C_{b_1}^e(2,1) \end{bmatrix} \quad (6)$$

It is important to mention that the later equation is approximate and just holds for small time steps $\Delta t$ and in case of a few larger intervals it induces a scale on the rotation rate which can be compensated by:

$$sc = \cos^{-1}\left(\frac{\Delta C_{b_1}^e(1,1) + \Delta C_{b_1}^e(2,2) + \Delta C_{b_1}^e(3,3) - 1}{2}\right) \quad (7)$$

$$\omega_{eb_1}^{b_1} = \frac{sc}{\sin(sc)} \omega_{eb_1}^{b_1} \quad (8)$$

Since in the proposed gimbal model, there is three nested mounted frames $g_T$, $g_R$, $b_I$ before the state is reached to the IMU frame, the set of equations (1) to (8) must be successively applied three times with generic frames ($b_1, b_2$) substituted by ($b_c \equiv g_P, g_T$), ($g_T, g_R$) and ($g_R, b_I$). This will eventually result in ground-truth PVA ($p_{b_I}$, $v_{b_I}^n$, $\Theta_{b_I}^n$) of the IMU body which is fed into kinematic equations.

### 3.3. Kinematics Equations

Having simulated ground-truth PVA ($p_{b_I}$, $v_{b_I}^n$, $\Theta_{b_I}^n$) of IMU body and their memorized values from previous time step, the goal is to compute the simulated ground-truth average specific force and rotation rate ($f_{ib_I}^{b_I}$, $\omega_{ib_I}^{b_I}$) of the IMU body relative to the inertial frame, resolved in IMU body frame. This is done according to kinematic equations (in ecef frame e) [1]:

$$\begin{aligned} \Delta C_e^i &= C_e^{i\ T}(t) C_e^i(t - \Delta t) \\ &= \begin{bmatrix} \cos(\omega_{ie}\Delta t) & \sin(\omega_{ie}\Delta t) & 0 \\ -\sin(\omega_{ie}\Delta t) & \cos(\omega_{ie}\Delta t) & 0 \\ 0 & 0 & 1 \end{bmatrix} \end{aligned} \quad (9)$$

where $\omega_{ie}$ is the earth rotation rate around its polar axis and $\Delta C_e^i$ is the delta CTM of earth rotation during $\Delta t$ time step. Using $\Delta C_e^i$ and old $C_{b_I}^e(t - \Delta t)$ and current $C_{b_I}^e(t)$ CTMs of the IMU body frame $b_I$ relative to the ecef frame e, the delta CTM $\Delta C_{b_I}^i$ of $b_I$ relative to the inertial frame i, is calculated as:

$$\Delta C_{b_I}^i = C_{b_I}^{e\ T}(t)\ \Delta C_e^i\ C_{b_I}^e(t - \Delta t). \quad (10)$$

To obtain ground-truth output average rotation rate signal $\omega_{ib_I}^{b_I}$ from delta CTM $\Delta C_{b_I}^i$ one can adopt equations (6) to (8) substituting scripts (e, $b_1$) by (i, $b_I$). The ground truth average specific force $f_{ib_I}^e$ in the ecef frame e is calculated from current $v_{eb_I}^e(t)$ and old $v_{eb_I}^e(t - \Delta t)$ velocity of the IMU body frame $b_I$ as follows:

$$\begin{aligned} f_{ib_I}^e &= \frac{v_{eb_I}^e(t) - v_{eb_I}^e(t - \Delta t)}{\Delta t} \\ &\quad - g^e\left(p_{b_I}(t - \Delta t)\right) \\ &\quad + 2\Omega_{ie}^e v_{eb_I}^e(t - \Delta t) \end{aligned} \quad (11)$$

where $g^e(.)$ is the earth gravitational model that computes $g$ vector in ecef frame e for the given input location and the skew symmetric matrix $\Omega_{ie}^e$ is defined as:

$$\Omega_{ie}^e = \begin{bmatrix} 0 & -\omega_{ie} & 0 \\ \omega_{ie} & 0 & 0 \\ 0 & 0 & 0 \end{bmatrix} \quad (12)$$

Finally the ground-truth specific force $f_{ib_I}^{b_I}$ resolved in the IMU body frame is calculated as:

$$f_{ib_I}^{b_I} = \bar{C}_{b_I}^{e\ -1} f_{ib_I}^e \quad (13)$$

where the rotation average $\bar{C}_{b_I}^e$ between two time steps (i.e. at $t - \frac{\Delta t}{2}$) is computed as follows:

$$\bar{C}_{b_I}^e = \begin{cases} C_{b_I}^e(t - \Delta t) * C_{b_I^-}^{b_I^-} - \frac{\Delta t}{2} \Omega_{ie}^e C_{b_I}^e(t - \Delta t) & |\alpha| > \varepsilon \\ C_{b_I}^e(t - \Delta t) - \frac{\Delta t}{2} \Omega_{ie}^e C_{b_I}^e(t - \Delta t) & o.w. \end{cases} \quad (14)$$

$$C_{b_I}^{b_I-} = I_3 + \frac{1-\cos(|\boldsymbol{\alpha}|)}{|\boldsymbol{\alpha}|^2} A_{ib_I}^{b_I} + \frac{1}{|\boldsymbol{\alpha}|^2}\left[1 - \frac{\sin(|\boldsymbol{\alpha}|)}{|\boldsymbol{\alpha}|}\right] A_{ib_I}^{b_I\ 2} \quad (15)$$

$$A_{ib_I}^{b_I} = \begin{bmatrix} 0 & -\boldsymbol{\alpha}(3) & \boldsymbol{\alpha}(2) \\ \boldsymbol{\alpha}(3) & 0 & -\boldsymbol{\alpha}(1) \\ -\boldsymbol{\alpha}(2) & \boldsymbol{\alpha}(1) & 0 \end{bmatrix} \quad (16)$$

where $\boldsymbol{\alpha} = \boldsymbol{\omega}_{ib_I}^{b_I} \Delta t$ is delta rotation vector of IMU body frame $b_I$ during the time step, $A_{ib_I}^{b_I}$ is its skew symmetric matrix and the $\varepsilon$ is a small positive constant, in orders of $10^{-8}$ to avoid division by zero and error magnification, caused by finite numeric precision.

### 3.4. IMU Error Modeling
The goal of IMU error models is to impose deterministic and stochastic disturbing factors on the ground truth average specific force and rotation rates ($\boldsymbol{f}_{ib_I}^{b_I}, \boldsymbol{\omega}_{ib_I}^{b_I}$) to model the final IMU simulated signals ($\tilde{\boldsymbol{f}}_{ib_I}^{b_I}, \tilde{\boldsymbol{\omega}}_{ib_I}^{b_I}$). This is commonly implemented using the following equation [1]:

$$\tilde{\boldsymbol{\omega}}_{ib_I}^{b_I} = \boldsymbol{b}_g + (I_3 + M_g)\boldsymbol{\omega}_{ib_I}^{b_I} + G_g \boldsymbol{f}_{ib_I}^{b_I} + \boldsymbol{w}_g \quad (17)$$

$$\tilde{\boldsymbol{f}}_{ib_I}^{b_I} = \boldsymbol{b}_a + (I_3 + M_a)\boldsymbol{f}_{ib_I}^{b_I} + \boldsymbol{w}_a \quad (18)$$

where the IMU biases, scale and cross coupling factors, noises and g-dependent biases and their unit is defined in Table 2.

Table 2: IMU error model parameters definition

| Parameter | Value | Unit |
|---|---|---|
| $\boldsymbol{b}_a$ | [0.009, -0.013, 0.008] | $m/s^2$ |
| $\boldsymbol{b}_g$ | [-0.175, 0.252, 0.155] x1e-3 | $rad/s$ |
| $M_a$ | rand 3x3 matrix x5e-3 | - |
| $M_g$ | rand 3x3 upper triangular x3e-3 | - |
| $\boldsymbol{w}_a$ | 7.845e-4 | $m/s^{3/2}$ |
| $\boldsymbol{w}_g$ | 2.327e-6 | $rad/s^{1/2}$ |
| $G_g$ | rand 3x3 matrix x1e-5 | $rad.s/m$ |

For real sensors, these parameters are usually specified in the datasheets. To simulate a specific sensor one can substitute the values from the corresponding data sheets.

## 4. Simulation Results and Evaluation
We have implemented the proposed gimbal mounted IMU simulator in MATLAB. The flight data simulated by JSBSim flight simulator using a Censa 310 (C310) airplane is captured by reading UDP packets. Following that, gimbal simulation, kinematic equations and finally IMU error simulations are computed consequitively as described in section 3. The numeric values for geometric parameters of the simulated gimbal (see Fig. 4) is listed in Table 3.

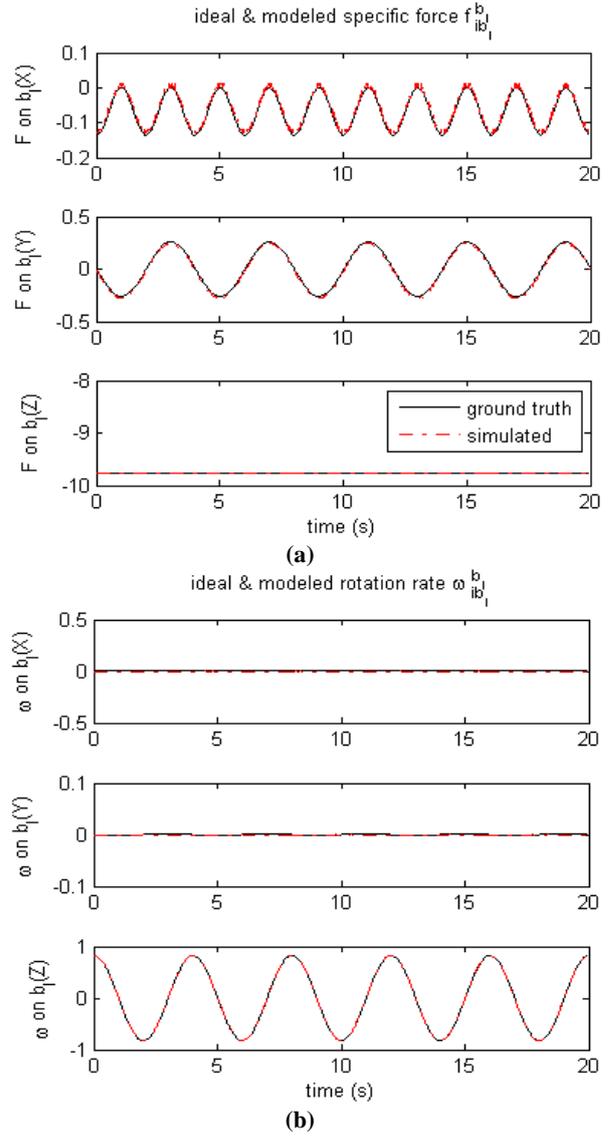

**Fig. 5.** Unit test of IMU signals applying a periodic *pan* rotation to gimbal considering a stationary platform **(a)** measured specific force **(b)** measured rotation rates

Table 3: Gimbal model parameter values

| Parameter | Value/Description | Unit |
|---|---|---|
| $l_{PT}$ | 0.1 | $m$ |
| $l_{TR}$ | 0.1 | $m$ |
| $P_P$ | 4 (the pan rotation period) | Sec |
| $P_T$ | 6 (the tilt rotation period) | Sec |
| $P_R$ | 10 (the roll rotation period) | Sec |
| $\theta_P$ | $\pi/6$ (the pan alternating magnitude) | $rad$ |
| $\theta_T$ | $\pi/6$ (the tilt alternating magnitude) | $rad$ |
| $\theta_R$ | $\pi/12$ (the roll alternating magnitude) | $rad$ |

To validate the simulated gimbal mounted IMU signals, we have conducted two sets of experimets. The first set is unit test which just applies some periodic rotations on the gimbal PTR angles in stationary and flight motions and visualizes the resulting acceleration and rotation to verify that the valid IMU signals are generated, comparing with the ground truth signals.

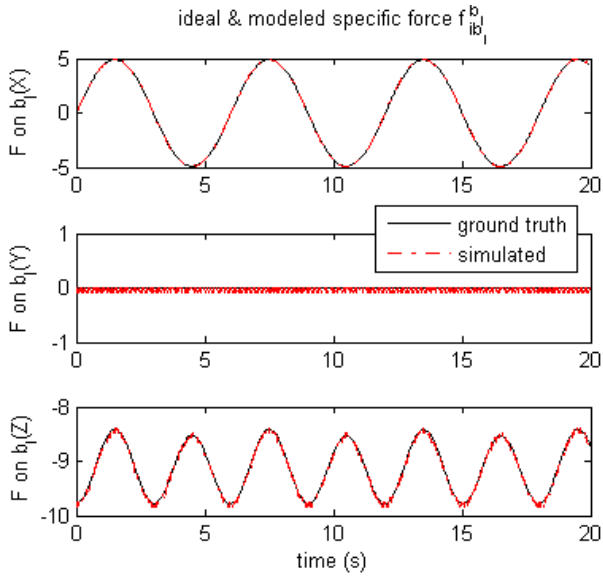

**Fig. 6.** Unit test of IMU signals applying a periodic *tilt* rotation to gimbal considering a stationary platform
**(a)** measured specific force **(b)** measured rotation rates

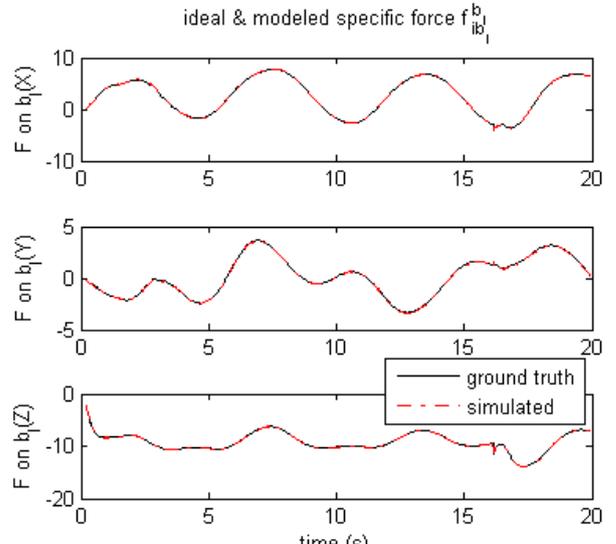

**Fig. 7.** Unit test of IMU signals applying joint periodic *pan*, *tilt* & *roll* rotation to gimbal during aircraft take off.
**(a)** measured specific force **(b)** measured rotation rates

Results for this set of experiments are illustrated in the Fig. 5 to 7. In Fig. 5, 6 the platform is stationary and isolated pan and tilt sinusoidal rotations with the period and magnitude listed in Table 2, are applied to gimbal servos respectively. The gimbal pan and tilt rotation offsets are set to zero in all cases to align its offset body frame $(x_{b_I}, y_{b_I}, z_{b_I})$ parallel to the platform body frame $(x_{b_c}, y_{b_c}, z_{b_c})$. This clears the situation to be tracked in mind and allows for verifying that true acceleration and rotation rate signal shapes are simulated by checking for the IMU axes that should expect some accelerations and rotation rates.

In Fig. 7 joint pan, tilt and roll sinusoidal rotations around zero offsets are applied to the gimbal servos and the gimbal is considered to be on an airplane that is starting a take off using the JSBSim. Hence the major non-periodic acceleration and rotation changes is visible on the down axis of the IMU which can be attributed to the aircraft elevation.

In the second set of experiments we have conducted an end to end integration/functional test of the simulated gimbal mounted IMU by applying its simulated signals to a losely couple aided INS integration algorithm. This algorithm is provided as an open source package in MATLAB by [1] which is also used in [3] for validation of their proposed IMU simulator that is direclty connected to the aircraft body.

We have modified the codes to mixin our MATLAB implementation of the proposed IMU signal simulator functions. We have also considered a simulated hypothetic aiding sensor (discussed in section 2, see Fig. 2), with similar observation type (position and vlocity) and accuracy as a typical GNSS/GPS sensors, but with the difference that it is viable to mount it on the inner frame of the same gimbal, co-located with the simulated IMU. This is to avoid the unnecessary compilication of testing suit which otherwise required compensation of the dynamic or static lever arms.

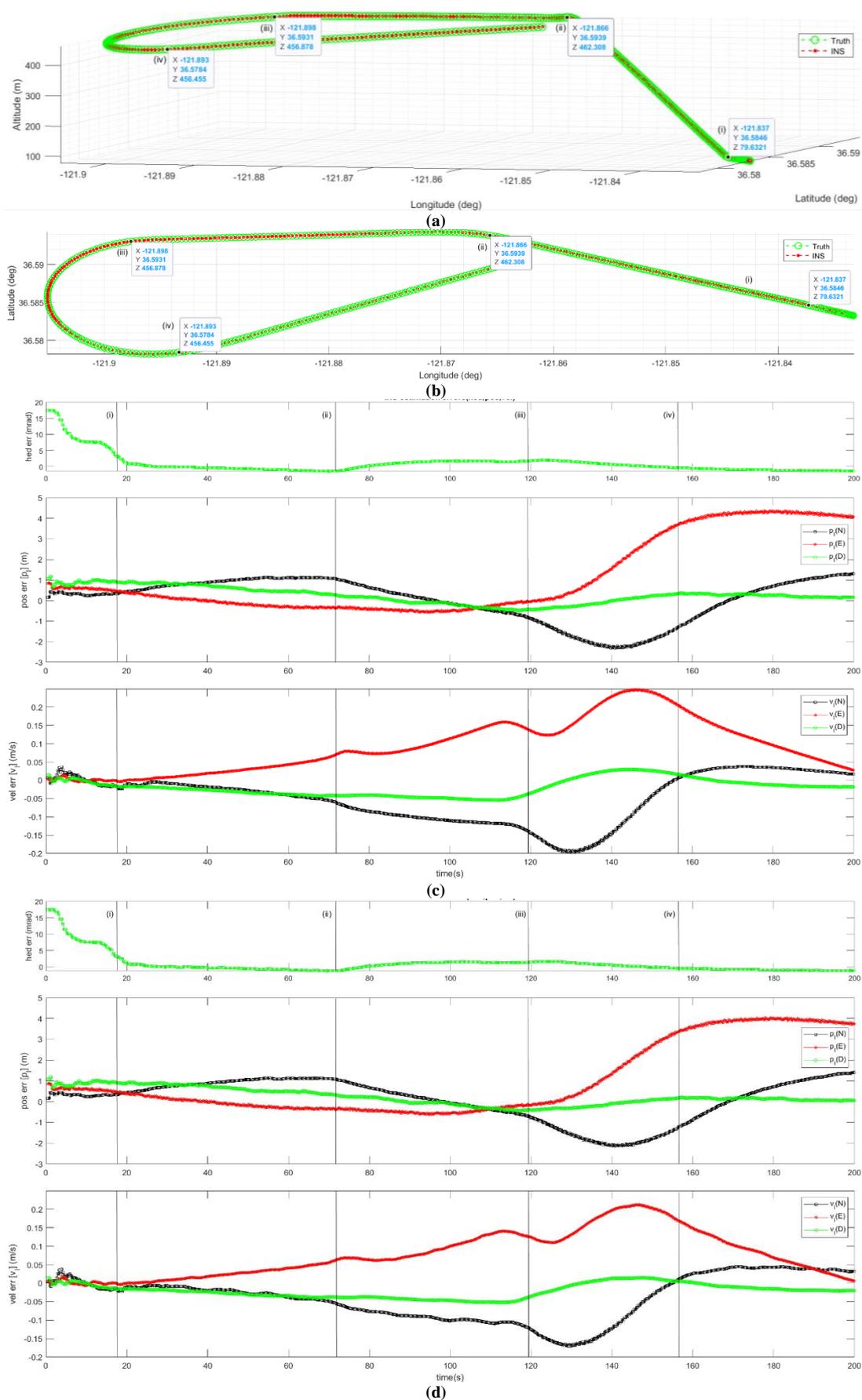

**Fig. 8.** Functional/Integration test of the proposed IMU simulator by running virtual-GNSS aided INS
(a) 3D flight way points (b) 2D view of flight way point (c) Position $p_{b_I}^n$ and velocity $v_{b_I}^n$ of IMU, mounted on aircraft body resolved in local ned coordinate (d) Position $p_{b_I}^n$ and velocity $v_{b_I}^n$ of IMU mounted on gimbal inner-frame resolved in local ned coordinate.
Marked flight points/instances: (i) start of ascending (ii) start of cruse (iii) start of strong turn maneuver (iv) end of trun.

The variable lever arm compensation is a requirement for the aided INS integration tasks but it is not necessary for unit/functional testing of the IMU simulator itself which is the goal of our evaluation experiment. Extended navigation results for 200 sec of flight simulation, involving various horizintal and vertical maneuvers are illustrated in the Fig. 8 . In this figure, in part (a) the 3D trajectory of aircraft's take off and fly is presented in LLA coordinates, in which the four time instances for abrupt changes in the motion are marked and in part (b) the 2D projection of the same trajectory from the top view is presented and in part (c) the navigation estimation errors when the simulated IMU is assumed to be mounted on the aircraft body frame is ploted and in part (d) the corresponding navigation estimation errors for the IMU being mounted on the inner farme of a gimbal that is itself onboard the airframe is ploted.

As results show, the errors for estimatioin of the heading $\boldsymbol{\Theta}_{b_I}^n(D)$, the position $\boldsymbol{p}_{b_I}^n$ and the velocity $\boldsymbol{v}_{b_I}^n$, resolved in the local ned coortinate for the platform-mounted and gimbal-mounted IMU simulated signals are in the same range where heading errors are about 3 mili-radians and position errors are about 2 meters and velocity errors are about 20 cm/s which is acceptable considering that the motion scenario is a take off with three drastic maneuvers.

From results of Fig. 8, it is evident that during the start of the maneuvers/motion changes (*i.e.* time instances marked by (i)-(iv)), the errors are initially increasing but the navigation filters manages to handle the situation after a while, so that the estimations is convergent toward the true values after a while.

For overal assessment of the navigation solutions the root mean square error (RMSE) of the heading, position and velocit estimations is reported in Table 4.

Table 4: Navigation estimation errors (RMSE)

| Gimbal Status | RMSE heading (mrad) | RMSE position (m) | | | RMSE velocity (m/s) | | |
|---|---|---|---|---|---|---|---|
| | $\boldsymbol{\Theta}_{b_I}^n(D)$ | $\boldsymbol{p}_{b_I}(N)$ | $\boldsymbol{p}_{b_I}(E)$ | $\boldsymbol{p}_{b_I}(D)$ | $\boldsymbol{v}_{b_I}(N)$ | $\boldsymbol{v}_{b_I}(E)$ | $\boldsymbol{v}_{b_I}(D)$ |
| OFF | 3.246 | 0.481 | 2.142 | 1.041 | 0.030 | 0.117 | 0.082 |
| ON | 3.176 | 0.473 | 1.969 | 1.002 | 0.028 | 0.099 | 0.073 |

Surprisingly we observe that the estimation errors for the case that IMU is mounted on the rotating frame of Gimbal is slightly lower than the cas that the IMU is considered to be fixed relative to the airframe body (by turning off the Gimbal. This can be justified by the improvement of observability of hidden variables provided by additional rotation of IMU which would provide more diverse measurements that increase the information needed for identification of the state variables.

## 5. Conclusion & Future Work

In this work we have proposed an IMU signal simulator mounted on a 3DoF Gimbal model onboard an airframe that exploits realistic flight dynamics model (FDM). Though there are IMU signal simulators for airborne platforms, considering FDM and also there exists INS systems which operate on signals coming from a real gimbal mounted IMU, but we could not find a 3DoF gimbal mounted IMU signal simulator that is additionally onboard an aircraft. The simulated IMU signals are validated by unit and integration tests.

Beside the simulated signals the simulator also provides ground truth position, velocity and attitude making it very helpful in development and comprehensive validation of variable lever arm, airborne aided SINS processing algorithms with gimbal mounted IMU. Such an extensive evaluation requires tests under many motion profiles scenario which is not feasible in real world specially in case of airborne platforms that need expensive flight logistics.

The goal of developing this gimbal mounted IMU signal simulator was two fold: (I) to enable comprehensive evaluation of existing commercial INS systems that feature a real gimbal mounted IMU. (II) to be used in our future work of developing losely and tightly coupled dynamic lever arm GNSS/INS integration algorithms with gimbal mounted IMU.

Commonly the gimbal is used as a stabilizer with a closed loop control operating on the raw gyro observations of the IMU mounted on it. Hence, it is interesting to simulate the gimbal dynamics model and the stabilization control loop as a more realistic stimuli for gimbal PTR angles (replacing current simulated semi periodic input rotations) as a future improvement.

## 6. Acknowledgement

Authors thank their beloved family members for their unwaivering support. This work is developed as part of an R&D project in the System Intelligizers Co. Research Dept. (SICO Research), Shiraz, Fars, Iran.